\documentclass[hf]{ceurart}

\usepackage{tikz}
\usepackage[english]{babel}

\usepackage[letterpaper,top=2cm,bottom=2cm,left=3cm,right=3cm,marginparwidth=1.75cm]{geometry}

\usepackage{amsmath}
\usepackage{graphicx}
\usepackage{multicol}
\newcommand{\rem}[1]{\relax}

\newlength{\mathfrwidth}
\newsavebox{\mathfrbox}
\newenvironment{mathframe}
    {\begin{lrbox}{\mathfrbox}\begin{minipage}{\mathfrwidth}}
    {\end{minipage}\end{lrbox}\noindent\fbox{\usebox{\mathfrbox}}}

\begin{document}

\copyrightyear{2023}
\copyrightclause{Copyright for this paper by its authors.
  Use permitted under Creative Commons License Attribution 4.0
  International (CC BY 4.0).}
\conference{14th MathUI Workshop 2023}  

\title{Extracting Mathematical Concepts \\
with Large Language Models}

\author[1]{Valeria de~Paiva}
[email=valeria@topos.institute]

\author[2]{Qiyue Gao}
[email=gaoq@rose-hulman.edu]

\author[3]{Pavel Kovalev}
[email=pakova@indiana.edu]

\author[4]{Lawrence S.~Moss}
[email=lmoss@indiana.edu]
\fnmark[1]
\fntext[1]{Supported by grant \#586136 from the Simons Foundation.}

\begin{keywords}
mathematical terminology \sep
knowledge graph \sep
annotation \sep
large language model 
\end{keywords}

\begin{abstract}
We extract mathematical concepts from mathematical text using generative large language models 
(LLMs) like ChatGPT, contributing to the field of automatic term extraction (ATE) and  mathematical text processing, and also to the study of LLMs themselves.   
Our work builds on that of others, 
in that we aim for automatic extraction of terms (keywords) in one mathematical field, category theory, using as a corpus the 755
abstracts from a snapshot of the online journal \emph{Theory and Applications of Categories}, circa 2020.  Where our study diverges
from previous work is in 
(1) providing a more thorough analysis of what makes mathematical term extraction a difficult problem to begin with;
(2) paying close attention to inter-annotator disagreements; 
(3) providing a set of guidelines which both human and machine annotators could use to standardize the extraction process;
(4) introducing a new annotation tool to help humans with ATE, applicable to any mathematical field and even beyond mathematics;
(5) using prompts to ChatGPT as part of the extraction process, and proposing best practices for such prompts;
and (6) raising the question of whether ChatGPT could be used as an annotator on the same level as human experts.
Our overall findings are that the matter of mathematical ATE is an interesting field which can benefit from participation by LLMs, 
but LLMs themselves cannot at this time surpass human performance on it.
\end{abstract}

\maketitle

\conference{14th MathUI Workshop 2023}

\section{Introduction}

This paper addresses an issue which must be confronted in large-scale user interaction with mathematical text, the isolation of mathematical concepts in text itself.   
We ask whether large language models (LLMs) can assist in the automatic extraction of terms from text.
For example, could one input a textbook or research monograph and ask an LLM to construct a reasonable index for it?
Obviously, if this were possible, then it would save authors time and energy.   
And if an LLM is of limited use, would it still be sensible to ask?
What would be the best way to prompt the LLM?


\subsection{Mathematical text and mathematical concepts}
Mathematical text is different from news or encyclopedic (Wikipedia-style) text. For a start, it has equations and diagrams and special fonts, usually laid out in \LaTeX.
It usually requires special processing. 
Mathematical text
employs highly specialized vocabulary; these are the topic of this paper.
It also adheres to special conventions that can put off beginners, but these make concepts and statements as clear and non-ambiguous as the author can possibly make them. 
Those conventions also come with their own terminology (see Section~\ref{subsection-math-writing}), but in this paper we make a distinction between
terms in the mathematical vocabulary and terms in the mathematical practice.   We are interested in automatically finding the
former rather than the latter.

The emergence of large language models (LLMs, e.g. GPT-3 \cite{gpt3}) has permanently altered the work of 
natural language processing (NLP) researchers.  It goes without saying that the general public has become aware of LLMs,
and over 100 million of people have used ChatGPT~\cite{ChatGPT}.  This is because
LLMs have  the capacity of learning downstream tasks given a few question-answer examples (\emph{in-context examples}) or even no example whatsoever
in the input. This mysterious and charming capacity motivates us to discover what role LLMs could play in our task of
\emph{extracting mathematical concepts from text}.  That is, we are not interested in using LLMs to prove theorems  
or to search the world's mathematical corpus for results.   We set our sights on something which at first glance 
looks ``smaller'' but which we have made a quick advance on.
We build a \emph{mathematical concept extraction pipeline} based on using ChatGPT.
We ask whether
if it serves as a good annotator, comparable to human experts, or 
whether instead it is
a decent helper that only produces preliminary results, in need of additional processing from human experts.

\paragraph{Contributions}
In this work we use ChatGPT~\cite{ChatGPT} to extract mathematical concepts\footnote{In this paper
we use the terms ``term'', ``concept'', and ``keyword'' interchangeably.
We know that there are situations where it is sensible to distinguish these terms,
but in the present paper we have no reason to do so.} from academic papers in Category Theory, a reasonably recent (mid-last-century) branch of pure mathematics.
ChatGPT is the only large language model which we use in this paper.  While the results of our work might well depend on the particular LLM used, as ChatGPT being one of the best performing models in the field,
we expect the main conclusions of our work would hold for other LLMs.
Using NLP tools for dealing with a specific domain such as mathematics is not very easy, as there is basically very little data annotated for mathematics. 
We make two contributions.
First, we analyze mathematical terms (Section~\ref{section-background}) and then propose guidelines
for human annotators (\ref{section-human-annotators}).
Second, we prompted ChatGPT for such terms (\ref{section-prompting}) and conducted three experiments
(\ref{section-experiments}) using a corpus (\ref{section-corpus}).
We report on these experiments and draw conclusions.
(\ref{section-conclusion})

\rem{
Experiment 1:  300 golden, then look at human agreement.
Take union, symmetric difference.
Mike Barr, 1989, andres kock wrong by ChatGPT
in the other way, internal preorder
But humans also are inconsistent on some of this.

421/523 in 3000 vcvp: NO, 421/523 in 300 sentences, not 3K sentences.


Exp 3: Four annotators!

}

\subsection{Related work}

 We have not been able to find much related work, especially, related data for the task we set ourselves. The only official NLP competition data we found, discussed in \cite{Luan2018, wadden2019}, was the corpus SciERC\footnote{http://nlp.cs.washington.edu/sciIE/}. This dataset  includes annotations for scientific entities, their relations, and coreference clusters for 500 scientific abstracts. (These abstracts are taken from 12 AI conference/workshop proceedings in four AI communities, from the Semantic Scholar Corpus.) SciERC extends previous datasets in scientific articles 
 related to the SemEval 2017 Task 10 and SemEval 2018 Task 7
 meetings
 by extending entity types, relation types, relation coverage, and adding cross-sentence relations using coreference links. However, as shown  by  \cite{collard2022, collard2023parmesan}, there seems to be a considerable difference between ``scientific entities" and ``mathematical entities" as the results reported by the software DyGIE++ of Wadden et al are surprisingly low in the mathematical text. Details of this corpus of annotated scientific entities can be found at \cite{scierc-corpus}.

Most of the work on extracting information from mathematical texts has concentrated on extracting meaning from formulas \cite{asakura2020}, which we do not consider. We want to see how much information we can obtain from mathematical vernacular alone, without symbols, formulas, or diagrams.

Another line of work that might be relevant is the work on ``flexiformalization". While we share the motivations and goals of  \cite{Kohlhase2014},  especially in believing that our extracted concepts should be mapped to an ontology such as WikiData, we do not strive for a formalization of the data model obtained from our concepts. 
{We are trying to work with existing mathematical text (category theory in this paper) using tools from contemporary NLP.   An important first step for us would be to construct 
mathematical datasets to be able to test our procedures, and this is what we do:
we construct several small datasets of mathematical text annotated for mathematical entities.}

\section{Background: difficulties with the isolation of mathematical concepts in texts}
\label{section-background}

The automatic extraction of mathematical terms was not as easy as we thought it would be.
Certainly for machines it turns out to be problematic.
For that matter, we have also found that human annotation of terms in 
mathematical texts comes with its own problems.
Our annotation has brought to the surface quite a number of issues concerning mathematical writing, and so we discuss these in this section.

\subsection{Words coming from mathematical writing that have little or no mathematical content}
\label{subsection-math-writing}

One persistent problem is that 
\emph{mathematical text contains quite a few words which are special to the genre of mathematical writing
but which do not themselves bear mathematical content}.
  Examples include 
 \emph{assumption}, 
 \emph{characterization}, 
 \emph{conjecture}, \emph{consequence},
  \emph{contradiction}, 
\emph{counter-example},
\emph{paper},
and
\emph{therefore}.
In some cases, the words are more generally part of academic writing beyond mathematics,
and they have different meanings in ``real life'' than in the academic world: \emph{paper} and
\emph{proposition},
for example.
We work with the assumption that such words would not be part of the index of mathematical terms in any book.
Yet if one asks a LLM for a list of words that are especially found in mathematical texts, the words above are likely to 
be found.

\subsection{Difficulties with alternate forms of words}

The ultimate goal of this work is the automatic extraction of a list of mathematical terms from text.
For this, it would be advantageous to avoid repeated entries, or even to have closely related entries.

\begin{enumerate}
\item \emph{Should nouns be listed in the singular or the plural, even if their appearance in the source text
is different?}  The difficulty here is not special to mathematics: it affects terminological indices in every subject whatsoever.
\item \emph{Should adjectives be listed alone, or only with the nouns they modify?} 
It would be simpler to drop the head noun, but this will not work.
Frequently adjectives have different meaning depending on the nouns they modify:
 \emph{regular polygon}, \emph{regular expression}, etc.
Even in our target subject of category theory, \emph{regular monomorphism} and \emph{regular epimorphism}
have different (but related) meanings.   This point also affects all disciplines.
\item \emph{If an adjective is inflectionally derived from a verb, should the automatic system prefer one form over another?}
For example, in \emph{interpolated function}, should an index show \emph{interpolate}, even though it might be missing
from the text?   Or should it show \emph{interpolated}, or \emph{interpolated function}, or even all of these?
\item \emph{In cases where there are constructions involving two or more adjectives handling a single noun,
what should the system do?}
If a text says \emph{differential graded category}, should we also add to an index sub-expressions like
\emph{differential category} and \emph{graded category}?  In many cases, the decision on this requires expert
knowledge (or perhaps access to much more data).   This situation is fairly common.  How should it be handled?
\item \emph{How should an automatic system handle expressions with prepositions?}
For example \emph{sheaf of germs of analytic functions}?
Should it include this expression, and also 
\emph{germs of analytic functions} and also \emph{analytic functions}?
Mathematical text is full of such expressions: \emph{area between the tangents
to two circles}, for example.
\item \emph{How should the system handle proper names, or upper-case symbols in the source data?}
In mathematical text, one does find proper names.  They might occur in adjective-like positions: as in \emph{Shanin's method},
\emph{Lagrange interpolation}, \emph{Birkhoff's Variety Theorem}.    In some cases, the name is part of the concept;
the concept would not be understandable without the name.  For example: a mathematical text which dropped \emph{Cauchy}
from \emph{Cauchy sequence} would likely have an error, and an index that omitted \emph{Cauchy sequence} in favor of \emph{sequence}
would not be helpful.
The same holds for \emph{Gauss' Lemma} and \emph{Turing machine}.
  Certainly \emph{Newton's Method} is a standard concept that books on calculus would want to index.
  In other cases, the proper name is much less standard.  Our data included \emph{Shanin's method}, and we ourselves do not know
  whether this is a standard usage or a nonce.

\end {enumerate}

\section{Human annotation and guidelines}
\label{section-human-annotators}
Looking simply at one or two examples this task of extracting concepts seemed easy enough. For a sentence like \textit{A notion of central importance in categorical topology is that of topological functor}, it was clear that we needed to extract the concepts of \textit{categorical topology} and  \textit{topological functor}. But it was also clear that one should know the concept of a \textit{functor} and maybe also the one of \textit{topology}. Thus, some sub-concepts of extracted concepts should also be concepts. But which ones?

We realized that we needed some guidelines for the extraction of concepts and decided to annotate a pilot set of a hundred sentences to determine the necessary guidelines to guarantee a good inter-annotator agreement between the human annotators. The set of these hundred sentences can be found at \url{https://github.com/vcvpaiva/NLIMath/blob/main/PilotTest100.txt}.


We formulate some generic guidelines for our annotations and list them in Figure~\ref{figure-guidelines}.

\begin{figure}[t]
\begin{mathframe}
\begin{itemize}
    \item Try to treat math concepts as black-boxes, as much as 
    possible. We do not expect users
    to know their full meaning. We intend to annotate any term that we think a mathematician might not know
    and might want to check in some glossary or  in some index.
    \item Use the singular, instead of the plural, for concepts. Thus, annotators are expected to extract from the sentence,  \textit{We show that both approaches give equivalent bicategories} the concept \emph{equivalent bicategory}.
    \item A generic decision was that terms like \textit{theorem, corollary, conjecture, theory} are very used in mathematics, but are not the kind of mathematical concept we want to extract from texts. These are almost meta-concepts.
    \item Even if names of important mathematicians sometimes appear in book indices, the mathematicians themselves are not mathematical concepts. Thus while \textit{Grothendieck's construction} is a concept, \textit{Grothendieck} by itself is not a mathematical concept.
    \item If one has a long span that is a concept, e.g. 
    \emph{enriched accessible categories}, we should also list the sensible sub-spans like 
    \emph{accessible category}.
    \item If a sentence contains a mathematical adjective together with a mathematical
    noun (e.g. \emph{nilpotent algebra}), then we include the adjective together with the noun, but not the standalone adjective. (And we also include the noun itself since we want all smaller spans.)
If a sentence contains a  mathematical  adjective without a  mathematical  noun (e.g. \emph{nilpotent case}), 
then we only include the standalone adjective.
If a sentence contains a mathematical adjective without any noun, we include the adjective.
    \item Try as much as possible to avoid prepositions occurring inside of concepts.
    In an example like \textit{sheaf of germs of analytic functions} we want \textit{sheaf},
    \emph{germ}, and \emph{analytic function}, but we don't want \emph{germs of analytic function}
    and we don't want the original (long) span.  That is, these complex concepts are something that a
    reader would naturally understand but not find in an index.
\end{itemize}
\end{mathframe}
\caption{Annotation guidelines.\label{figure-guidelines}}
\end{figure}

However, while annotating carefully the initial sentences we noticed that these generic guidelines were not enough. 
Some further guidelines for annotation were created from this pilot annotation exercise. 
\begin{enumerate}
\item Certain words and expressions can be either math concepts or can be used in their English sense, for example 
\emph{method} in \emph{Shanin's method}. If the word is used as a mathematical concept, we want to use the whole expression. So for example in \emph{abelian group}, the word \emph{group} is used in the mathematical sense, so the concept is the whole expression. But if the word is used as a common English noun we do not want it as part of the concept. This is one of the reasons for disagreements between annotators, is \emph{method} above used as common English word or as mathematical concept? Similarly, for \emph{2-categorical refinement}, is this refinement a mathematical concept?

\item We believe we should do `coreference resolution' for our terms. 
So for the sentence \textit{We introduce a new class of categories generalizing locally presentable ones} we want to extract the term \textit{locally presentable category}.

\item For a long mathematical expression like \emph{skew monoidal categories}, 
suppose that the annotator is not sure of its meaning.
Suppose also that some sub-strings do make sense (monoidal category), while others might not e.g. 
\emph{skew category}. Unless we know the math involved, we cannot tell all the sub-concepts of a long expression. So this is a known reason for disagreement between annotators.    Even though the annotators
are instructed to set aside their mathematical training, this is frequently too much to ask.

\item The hardest piece of guideline is what to do about adjectives like 
\emph{analytic}, \emph{algebraic} or \emph{categorical} that are not concepts themselves, but that might indicate, the existence of a concept, in say \emph{2-categorical refinement} above.
\end{enumerate}

Then we asked ChatGPT to find concepts following the theme of our guidelines above and compared results obtained by 
ChatGPT with the ones from our human annotators.

\section{Corpus preparation}
\label{section-corpus}

We use an already prepared corpus, consisting of 755 abstracts from the open source journal \textit{Theory and Application of Categories}\footnote{\url{http://www.tac.mta.ca/tac/}}, processed around 2020. This small corpus (around 3.2K sentences) is available from GitHub\footnote{\url{https://github.com/ToposInstitute/tac-corpus}}. The abstracts were processed with the BERT version of spaCy\footnote{\url{https://spacy.io/}}. The corpus was already used before for extracting concepts using NLP tools~\cite{collard2022}. Here we are exploring what generative models, especially ChatGPT, can help with.

Since we are dealing with abstracts, we have less \LaTeX{} markup than if we were using full papers. We also have no diagrams, and spaCy provides us with lists of nouns, proper nouns, compounds, and adjective-noun terms, following the conventions of Universal Dependencies~\cite{nivre2020}.
Long sentences are likely to complicate the parsing and very short ones can be parsing or sentencization errors, so we use an extracted collection of sentences with moderate size and free from LaTeX  mark-up.
These can be found at \url{https://bit.ly/tac-examples}. This file also has examples of humanly-extracted concepts for each sentence.
Using this collection of sentences, a
set of humanly-extracted mathematical concepts was created for a random set of 436 sentences from 
the TAC abstracts.
We used three annotators from our team.
They could check the context of the sentences, if desired. Then we wanted to measure the extent of agreement between 
these three annotators.  This turned out to be more complicated than expected.

\subsection{New annotation tool}
\label{section-annotation-tool}

To discover true disagreements and reduce the noise produced by minor mistakes, we have repurposed a tool from previous work of Chen, Gao and Moss~\cite{chen-etal-2021-neurallog} to help with the human annotation. Figure \ref{fig:platform_overview} in Appendix \ref{sec:platform} gives an overview of this annotation platform. After uploading the dataset in the `upload' tab, users can select the dataset in a drop-down menu and specify a starting index to initiate the annotation process. One can then select consecutive words from the provided sentence, and they will be entered into the textbox below, where further edits could be done such as changing from plural to singular forms. Each annotated concept is listed under the sentence and can be removed, if necessary. These annotations are stored in our database upon submission. To retrieve the annotations for a specific dataset, users can use the `download' button at the bottom of the platform.

\section{Prompting ChatGPT}
\label{section-prompting}

\begin{table}[ht]
\begin{center}
\resizebox{\textwidth}{!}{
\begin{tabular}{l}

      \toprule
      Given the following Context, extract the words that denote Math concepts. \\
    \\
    \hspace{5mm} Here are some examples: \\
    \hspace{5mm} \{\textit{in-context}\; \textit{example}\} \\ 
    \\
    Now please solve the following problem. \\
    \\
    Context: \{\textit{math\_sentence}\} \\
    Concepts:   \\
      \bottomrule
\end{tabular}
}
\caption{The initial prompt template, where \textit{math\_sentence} denotes the sentence from which we wish to extract math concepts.}
\label{tab:prompt1}
\end{center}
\end{table}

\begin{table}[ht]
\begin{center}
\resizebox{\textwidth}{!}{
\begin{tabular}{l}
      \toprule
      \\
      \\
Context: `Let PreOrd(C) be the category of internal preorders in an exact category C.'\\
Concepts: [`internal preorder', `exact category']\\
\\
      \bottomrule

\end{tabular}
}
\caption{An example of the in-context demonstration.}
\label{tab:example1}
\end{center}
\end{table}

\begin{table}[ht]
\begin{center}
\resizebox{\textwidth}{!}{
\begin{tabular}{l}

      \toprule
      Given the following Context, extract the words that denote Math concepts. \\
    \hspace{5mm} \textcolor{red}{Be sure to make the concept words singular. For example when we see `functors' in a sentence,} \\
    \hspace{5mm} \textcolor{red}{we would extract `functor' rather than `functors' or when we see `categories'}\\
    \hspace{5mm} \textcolor{red}{we would like to extract `category' instead of `categories'!} \\

    \\
    Here are some examples: \\
    \{\textit{in-context}\; \textit{example}\} \\ 

    \\
\hspace{5mm} \textcolor{orange}{Also note that we are looking for concepts like modulation, enriched orthogonality, holonomy,} \\
\hspace{5mm} \textcolor{orange}{localization, variety, but not words shown in daily English sentences like `future work',}\\
\hspace{5mm} \textcolor{orange}{`conclusion', `this property'!}\\
\\
\hspace{5mm} \textcolor{blue}{We don't want a person's name to be extracted as a math concept although we understand that} \\
\hspace{5mm} \textcolor{blue}{a person's name could be part of the phrase that denotes a math concept.}\\
    \\
    Now please solve the following problem. \\
    \\
    Context: \{\textit{math\_sentence}\} \\
    Concepts:  \\
      \bottomrule
\end{tabular}
}
\caption{The updated prompt template with specialized instructions shown in different colors.}
\label{tab:prompt2}
\end{center}
\end{table}

\begin{table}[ht]
\begin{center}
\resizebox{\textwidth}{!}{
\begin{tabular}{l}
      \toprule
      \\
Context: `If the category is additive, we define a sheaf of categories of analytic functions.'\\
Concepts: [`additive category', `sheaf', `analytic function']\\
\\
Reason: the concept `additive category' is generalized from the sentence because of the original phrase `category is additive';\\ 
the concept `sheaf' is a known math concept shown in the sentence;\\
the concept `analytic function' is extracted from the original phrase `analytic functions' by removing the plural form;\\
here we don't want the single `category' and `additive' extracted
as math concepts since they are usual words \\
nor do we want the phrase `category of analytic functions' since the more concise phrase `analytic function' \\
should be extracted instead as a math concept.\\
      \bottomrule
\end{tabular}
}
\caption{An example of the \textbf{detailed} in-context demonstration.}
\label{tab:example2}
\end{center}
\end{table}

%

Table \ref{tab:prompt1} shows the initial prompt given to ChatGPT. The prompt starts with a concise description of the task followed by several in-context examples. An example of in-context demonstration is shown in Table \ref{tab:example1}. After these demonstrations, the real problem is given and the model is asked to extract concepts from this context. We first conducted a pilot study using this prompt on the random set of 300 sentences from the TAC abstracts. The evaluation from our human experts shows that the concepts generated from each sentence share common mistakes, including keeping the plural form as it appears 
in the context, generating a person's name and identifying everyday non-mathematical
used words as concepts. To help the model avoid such mistakes, we added specialized instructions in the prompt, as shown in Table \ref{tab:prompt2}.
These instructions alleviate the above issues but do not
solve them, since the model failed to generalize to words not mentioned in these instructions. We find common words like ``example'' have been extracted, and plural words like ``measures" not been transformed to
the singular. We think that detailed explanations of why 
some words are not mathematical concepts might help the model to understand our criteria and the categorization of mathematical concepts. This motivates us to construct detailed in-context examples(e.g. Table \ref{tab:example2}) that constitutes our final prompt, where ordinary in-context examples are replaced by a few detailed ones.



\section{Experiments and evaluation}
\label{section-experiments}

We have carried out three experiments with human and machine annotators.
\begin{enumerate}
    \item An in-depth pilot with 100 sentences from the TAC corpus.
    We aimed for sentences with little or no special symbols and which were of moderate length.
    \item A longer study with sentences from the same corpus.
     We chose 436 sentences from TAC abstracts (see \url{bit.ly/3DKEitc}). This file also has examples of humanly-extracted concepts for each sentence. 
For this longer set we use an industry strategy: We compare one human annotator with ChatGPT for all terms extracted. If they disagree, a second human adjudicator is invoked and their decision is final. 

    \item A full-scale attempt to use ChatGPT to annotate 55K sentences from the nLab website.
 We discuss this in Section~\ref{section-55K}.   
\end{enumerate}

\subsection{First experiment: pilot with 100 sentences from the TAC corpus}

As mentioned, our first experiment was based on 100 sentences from the TAC corpus.
Here are examples of sentences from this study:

\medskip

\begin{mathframe}
\begin{narrower}
We check these extra assumptions in several categories with pretopologies.

Functors between groupoids may be localised at equivalences in two ways.

We show that both approaches give equivalent bicategories.

In this paper, we use the language of operads to study open dynamical systems.

The syntactic architecture of such interconnections is encoded using the visual language of wiring diagrams.

\end{narrower}
\end{mathframe}

\medskip

\noindent
Three members of our team annotated 100 of these sentences, and at the same time ChatGPT was prompted.  
The annotators were given a spreadsheet
with the sentences in a single column, and next to this column were empty columns in which they were
asked to list all of the terms that should appear in a mathematical knowledge graph or an index
of terms in category theory.   At first
  many `minor mistakes' were made by the annotators. They  introduced typos, empty spaces and forgot to reduce concepts to singular terms, for instance. There were also many cases of annotators forgetting to read part of a sentence or treating some mathematical terms  as common English nouns.

It was for this reason that we formulated the guidelines mentioned in Section~\ref{section-human-annotators}.
This, too, was not a straightforward matter.  There were differences at every step.  Guidelines 
coordinate the annotation practice of the team members, but occasionally they sacrifice 
completeness in the process. 
 
In order to make the annotation process easier and more reliable, we used the tool mentioned in 
Section~\ref{section-annotation-tool}.

Here are our results on annotation of 100 sentences from the corpus, after filtering.
Out of 327 concepts extracted by the 4 annotators (three humans and ChatGPT) from the 100 sentences in the pilot experiment,  we have 120 concepts
that all four annotators agree on.  This gives (only) 37\% full agreement. 
This is pre-filtering. After filtering and some light changes (removing plurals and some common nouns like ``decade", etc.)  we get about 40\% full agreement.  This still seems low as a measure of full agreement.

\begin{table}[ht]
\centering
\begin{tabular}{lc}
\toprule
 Annotator & Number of Concepts \\
\midrule
 Annotator 1 & 206 \\
 Annotator 2 & 199 \\
 Annotator 3 & 194 \\
  ChatGPT (pre-filtering) & 235 \\
 The Union of All (pre-filtering) & 327 \\
 ChatGPT (post-filtering) & 226 \\
 The Union of All (post-filtering) & 310 \\
\bottomrule
\end{tabular}
\end{table}

There are 40 concepts that the three human annotators agree are  concepts, but ChatGPT does not extract (about 12\% of the total). Some examples include \textit{internal presheaf, gerbe} that only appear in the plural. Many are terms that human annotators considered important subspans of terms ChatGPT considers mathematical concepts. Examples include \textit{triple category, probability distribution, topology}, respectively, subspans from terms \textit{strict triple category, joint probability distributions, categorical topology}. 
See Figure~\ref{figure-discrepancies} for further examples.

It is not so surprising that
 ChatGPT did not extract all the concepts we want.
 What seems really surprising is the lack of agreement between the human annotators. 
Especially because the lack of agreement does not look conceptual, but simply noise in the extraction process. But how much is conceptual, how much is noise or minor mistakes? 

In addition, ChatGPT extracted concepts that are unwanted: again,
see Figure~\ref{figure-discrepancies}.
The one member of our team who did not annotate then \emph{filtered} 
 the terms extracted by ChatGPT.
They looked at the list of concepts that appeared in the ChatGPT list but not in the human list and deleted items that they considered to not be mathematical concepts; a few items were also added.

\label{section-results-of-interest}

\begin{figure}[t]
\begin{mathframe}

\medskip

Here are some of the concepts which all three human annotators found but ChatGPT missed:

\medskip

\begin{tabular}{l}
Grothendieck's \\
 \qquad six operations\\
Lie algebra\\
arithmetic variety\\
cartesian closed category\\
closed category\\
graph rewriting\\
\end{tabular}
\
\begin{tabular}{l}
group\\
homotopy\\
left proper model structure\\
morphism axiom\\
pointed regular\\
\qquad protomodular category\\
probability distribution\\
\end{tabular}
\
\begin{tabular}{l}
quotient triangulated \\
\qquad category\\
representation\\
smooth stack\\
sup-lattice\\
topology\\
triple category\\
\end{tabular}

\medskip

\hrule

\medskip

Here are some of the concepts found by ChatGPT which none of the three human annotators found:

\medskip

\begin{tabular}{l}
Grothendieck\\
acyclic models method\\
algebraic context\\
analysis\\
application\\
approach\\
axiom\\
balanced category\\
calculate\\
categorical\\
\end{tabular}
\quad
\begin{tabular}{l}
categorical property\\
category theory\\
characterization\\
closed monoidal\\
consequence\\
construction\\
corollary\\
definition\\
example\\
issue\\
\end{tabular}
\begin{tabular}{l}

language\\
localize\\
locally presentable\\
mathematically natural\\
motivation\\
non-abelian\\
open question\\
perspective\\
previous work\\
proof\\
\end{tabular}
\begin{tabular}{l}
property\\
prove\\
result\\
six operation\\
structure\\
symmetric monoidal\\
tame\\
theory\\
trivial\\
uniqueness statement\\
\end{tabular}
\end{mathframe}
\caption{Results discussed in Section~\ref{section-results-of-interest}.\label{figure-discrepancies}}
\end{figure}

\subsection{Evaluating the pilot}

The evaluation strategy uses a Jupyter notebook to transform the chosen terms into zeros and ones to calculate agreement.
We measure agreement using the \emph{Jaccard similarity index}.
For two sets $A$ and $B$, the definition is
\[
J(A,B) = \frac{|A\cap B|}{|A \cup B|}.
\]
We are given lists of terms,  one for each annotator (three humans and one non-human).
We produced one ``master-list'' that contains all concepts listed by at least one annotator (with no repetitions). We then created an empty dictionary with the purpose of storing the counts of each unique concept for each annotator. By iterating over each of the four lists and utilizing the master-list, we populated the dictionary with the desired value counts for each unique concept and converted it into a dataframe where the columns correspond to the annotators and the rows correspond to the unique concepts from the master-list. Since each concept from the master-list is present at most once in each of the four individual lists of concepts, the possible value counts are either 0 or 1. The entry at the intersection of row $i$ and column $j$ is 1 if and only if annotator $j$ listed concept $i$.

\rem{
This approach makes it easy to detect (dis)agreements. For example, to find all concepts that were listed by the human annotators but not by ChatGPT, it is enough to filter the dataframe by the condition that the sum of the entries in the three columns corresponding to the human annotators equals three and the entry in the remaining column that corresponds to ChatGPT equals zero.
}

Figure~\ref{Fig-experiment-1} presents
a set of comparisons of the  Jaccard similarity indices.
The main message is that by agreeing to guidelines, the human annotators
were able to achieve pairwise similarity scores in the range $0.75-0.8$.
They were not able to go higher.   On the other hand, 
ChatGPT's Jaccard index relative to humans is about $0.5$ even with filtering.

\begin{figure}[t]
\centering
\rem{
\begin{tabular}{ll}
\toprule
 Annotators Being Compared & Jaccard Score \\
\midrule
 annotator 1 and annotator 2    &  $0.753$ \\
 annotator 1 and annotator 3    & $0.794$ \\
 annotator 2 and annotator 3    &$0.746$ \\
\bottomrule
\end{tabular}
\rem}
\begin{tabular}{ll}
\toprule
 Annotators Being Compared & Jaccard Score \\
 \midrule
  annotator 1 and annotator 2    &  $0.753$ \\
 annotator 1 and annotator 3    & $0.794$ \\
 annotator 2 and annotator 3    &$0.746$ \\
\midrule 
\midrule
 ChatGPT and annotator 1        &  $0.485$ \\
 ChatGPT and annotator 2        & $0.518$ \\
 ChatGPT and annotator 3        &$0.505$ \\
\midrule
 ChatGPT and union of the humans & $0.45$ \\
 ChatGPT (after filtering) and union of the humans &   $0.5$ \\
\bottomrule
\end{tabular}
\caption{Comparison of Jaccard similarities in experiment 1.\label{Fig-experiment-1}}
\end{figure}

\label{section-results}

\rem{
Annotator 1 found 206.

Annotator 2 found 199.

Annotator 3 found 194.

ChatGPT found 226.

The union of 4 found 310.
}


\rem{
\begin{verbatim}
    

Humans = 0, GPT = 0: Length = 0
Humans = 0, GPT = 1: Length = 71
Humans = 1, GPT = 0: Length = 29
Humans = 1, GPT = 1: Length = 8
Humans = 2, GPT = 0: Length = 20
Humans = 2, GPT = 1: Length = 22
Humans = 3, GPT = 0: Length = 34
Humans = 3, GPT = 1: Length = 125

Humans < 1, GPT = 0: Length = 0
Humans < 1, GPT = 1: Length = 71
Humans < 2, GPT = 0: Length = 29
Humans < 2, GPT = 1: Length = 79
Humans < 3, GPT = 0: Length = 49
Humans < 3, GPT = 1: Length = 101

Humans > 0, GPT = 0: Length = 83
Humans > 0, GPT = 1: Length = 155
Humans > 1, GPT = 0: Length = 54
Humans > 1, GPT = 1: Length = 147
Humans > 2, GPT = 0: Length = 34
Humans > 2, GPT = 1: Length = 125
\end{verbatim}
}

\rem{
Pavel = 1
Valeria = 2
Larry = 3

Jaccard Similarity between Human (=union of 3 humans) and GPT (before filtering): 0.44954128440366975
Jaccard Similarity between Human (=union of 3 humans) and GPT (after GPT was filtered by Bert): 0.5
Jaccard Similarity Score between Human (=union of 3 humans) and GPT (after GPT was filtered by Bert and after Pavel removed junk from ChatGPT): 0.5849056603773585

Jaccard Similarity between Pavel and Valeria: 0.7532467532467533
Jaccard Similarity between Pavel and Larry: 0.7937219730941704
Jaccard Similarity between Larry and Valeria: 0.7466666666666667

Jaccard Similarity between Human and GPT: 0.5
Jaccard Similarity between GPT and Valeria: 0.5178571428571429
Jaccard Similarity between GPT and Pavel: 0.4845360824742268
Jaccard Similarity between GPT and Larry: 0.5053763440860215
}


\rem{
\newcommand{\veps}{x}
\newcommand{\blankcols}[1]{\multicolumn{#1}{c}{}}

\begin{center}
\setlength\extrarowheight{2pt} 
\setlength\arraycolsep{3pt} 
$\begin{array}{@{} r | *{8}{w{c}{9mm}|}}
\cline{1-2}
\mbox{ annotator 1}     & \veps & \blankcols{7} \\ 
\cline{1-3}
2     & \veps & 01 & \blankcols{6} \\
\cline{1-4}
3   &   & \veps & \veps & \blankcols{5} \\
\cline{1-5}
\mbox{ChatGPT}  & 0 & \veps & \veps & 0 & \blankcols{4} \\
\cline{1-6}
\rem{
0,1,5 & 1 & \veps & \veps & 1 & 0 & \blankcols{3} \\
\cline{1-7}
1,2   & \veps & & 01 & \veps & \veps & \veps & \blankcols{2} \\
\cline{1-8} 
0,3,4 & 0 & \veps & \veps & 0 & & 0 & \veps & \blankcols{1}\\
\hline
0     & \veps & 1 & 1 & \veps & \veps & \veps & 01 & \veps \\
\hline
 & 0 & 1 & 2 & 0,3 & 0,4 & 0,1,5 & 1,2 & 0,3,4 \\
}
\end{array}$
\end{center}
}

\subsection{Second experiment: 436 sentences from the TAC corpus}
As previously mentioned, in experiment 2, one human annotator and ChatGPT 
independently extracted concepts from 436 TAC sentences, and then a second human 
 \emph{adjudicated} conflicts between the human and ChatGPT.
 This same person then filtered ChatGPT's output, 
 as was done in experiment 1.
 Out of 250 items, 147 (or about 59\%) were deleted. It is worth noting that they were not very strict. For example, the adjudicator did not remove standalone adjectives
 (e.g. \emph{abelian} or \emph{accessible})
 even if they already appeared as part of larger math concepts in ChatGPT's list. This is more relaxed than our guidelines for humans, which say that standalone adjectives may only be included in certain cases.
 But it is not reasonable to expect that ChatGPT would be able to follow such a complicated guideline without being provided a variety of examples illustrating when it does and does not apply.

The adjudication-cum-filtering
process increased the Jaccard similarity index between the human annotator from 0.531 to 0.631. The adjudicator/filterer did not alter the other difference list -- the human list minus ChatGPT list -- mainly because humans are unlikely to include items that are obviously not mathematical concepts. There might be disagreement between the human annotator and adjudicator as to whether an item like \emph{Shanin's method} is indeed a concept, but settling such cases was not in the agenda of experiment 2. Instead, we wanted to only exclude items that are obviously not
mathematical concepts.  For example, 
\emph{previous work} is a concept found by ChatGPT.  But this is an error:
 no human being would think or say that this is a concept of mathematics.

\rem{
but what I meant that the same way that we looked at the concepts from the 100 sentences and thought no human could have imagined some as  concepts and discarded 17, I thought you had done the same for the 436 sentences. 
That is that  you looked at all the concepts that ChatGPT produced, and got rid of the outlandish ones that no human would have produced and we left the other ones that a human could have produced. 
and yes, I know that for the 55K ones we have no way of doing this initial filtering of concepts, but the thought is that if our live experiment works and we get more voluntary annotators we will get close to not needing this initial filtering. I also reckon that more prompt engineering will converge for better concepts.
 of course I am being optmist about chatGPT possibilities, but as I wrote earlier on, if we keep all the concepts that chatGPT has produced that we all think are ludicrous, we're shooting ourselves in the foot too much.

if we don't do this filtering our numbers are much worse, but this is fine, if you guys prefer not to present this discussion. I think it's worth writing about human-like mistakes and machine-only mistakes.
}

Here are  the results of this experiment.

\begin{table}[ht]
\centering
\begin{tabular}{ll}
\toprule
 Measurement & Value \\
\midrule
 Length of Human & 673 \\
 Length of ChatGPT & 740 \\
 Length of ChatGPT minus Human & 250 \\
 Length of Human minus ChatGPT & 183 \\
 Length of Adjudicated and Filtered ChatGPT minus Human & 103 \\
 Length of ChatGPT after Adjudication and Filtering & 593 \\
\bottomrule
\end{tabular}
\end{table}

\begin{table}[ht]
\centering
\begin{tabular}{ll}
\toprule
 Measurement & Jaccard Score \\
\midrule
 Between Human and Un-filtered ChatGPT & 0.531 \\
 Between Human and Filtered ChatGPT & 0.631 \\
\bottomrule
\end{tabular}
\end{table}

Experiment 2 had two main problems: 
The results depend too much on the humans used for the experiment; we would prefer to have more humans involved, to decrease the subjective aspects.
Second,  the more specialized the mathematics the harder it is for mathematicians to decide what is really a concept that it is worth keeping, versus what is an auxiliary definition 
that will not be adopted by others.  
From this perspective it would make more sense to work with text from more established mathematics, textbooks, where concepts have been crystallized by use, as opposed to journal articles, where the concepts are being forged. This leads us to our third experiment, where the text comes from a standard Category Theory resource, the wiki nLab.

\subsection{For the near future: terms extracted
from 55K sentences in a standard resource}
\label{section-55K}

A currently standard source for category theory at all levels is the nLab website
\url{https://ncatlab.org/}.    We have run ChatGPT with our prompts on a subset of
55K sentences from this website; see
\url{https://raw.githubusercontent.com/ToposInstitute/nlab-corpus/main/nlab_examples.csv}.
Of course we do not have human judgments to serve as a gold standard for these 55K sentences.
We do have ChatGPT's  concepts and we propose to use them as the automatically extracted concepts
(filtered to some degree),
given the relatively good results obtained in the two previous experiments.

Our aim is to put this information on a website as part of a 
large-scale effort to allow mathematicians to interact with our work.
Users will be able to approve or disapprove of ChatGPT's selection of
concepts.  We plan to use a modification of the same tool  which we mentioned in Section~\ref{section-annotation-tool}.
This ``living experiment" is somewhat similar to the first half of the experiment proposed in the blogpost ``Introducing the MathFoldr Project"\footnote{https://topos.site/blog/2021/07/introducing-the-mathfoldr-project/2020}. While on the earlier experiment we used NLP tools (then spaCy and Universal Dependencies) to produce the concepts that we wanted to map to WikiData, now we can use ChatGPT for the same purpose and hopefully the results are better.

Our task all along has been to ask
whether a given word
is or is not a ``mathematical concept''.  We do this by
asking: does the word belong
in a knowledge graph of the mathematical subject under discussion, 
and does it belong in an index?  Our work has shown that this question
is subtle and that there are many borderline cases.   We
plan to ask the community of category theorists who use platforms like Zulip for assistance.

This will have three results: first, it is likely to lead to additional sharpening of our understanding of mathematical terminology in the first place. (We are not aware of any individuals or groups looking at this topic from a data-driven perspective.)
Also, the involvement of the community will lead to a better appreciation of our topic, and of Mathematical User Interaction more broadly.
Thirdly, once in possession of a substantial number of concepts in several areas of mathematics we can organize this knowledge in structured ways, building a comprehensive knowledge graph. More importantly, the organized concepts should allow us to help with the formalization of mathematical results. There is a large movement towards formalizing mathematics using proof assistants nowadays. We believe that this kind of formalization will be more successful if we build it from the actual lingua franca of mathematicians, which is mathematical English.

\section{Conclusions}
\label{section-conclusion}

ChatGPT can help with the extraction of mathematical terms from mathematical texts, but at this time LLMs cannot replace humans.
With judicious prompting, ChatGPT
can find terms, but it misses some important ones that a mathematically-literate human would want, 
such as terms which are not fully explicit in a text but which a human would construct on the fly.
For example, in a sentence like \emph{We conjecture that this holds for all rings, including non-Artinian ones},
a reader would know that \emph{Artinian rings} belongs in a knowledge graph, even though it is not
explicit.  
An LLM might not extract that term.
At the same time it is likely to extract
terms that would not normally be considered ``mathematical terms,''
such as \emph{conjecture} or even \emph{ones}.

The fact that ChatGPT may be called almost effortlessly and gives acceptable preliminary
results suggests that it will be used in the future.  Our work gives numbers that serve as baselines
to future work in the area, both in terms of inner-annotator agreement by humans (measured by Jaccard index)
and also as in the quality of the LLMs findings.

We ourselves learned a lot about mathematical text by doing the annotations in a serious way.
Three people from our team have extensive experience with both mathematics
(even category theory)
and computational linguistics.
Two of us have spent years close-reading and correcting a corpus of natural language inference.
Yet we found it surprising that it was difficult to agree on borderline cases, such as
multi-word expressions containing adjectives or prepositions.  Returning to our example above,
is \emph{non-Artinian} a concept one would want in a knowledge graph or index?  We think not.
But again, it has proved too difficult to characterize the set of terms that one would want.
We believe that our guidelines for annotators are a good first step in such a characterization.

\rem{
\section{The concepts found from the 3000 sentences}

We took the 755 abstracts from the journal \emph{Theory and Applications of Categories} (TAC) and then chose 3000 sentences from these abstracts.
Here are some of the abstracts.

Then we asked ChatGPT to find keywords.
We used the following prompt.

ChatGPT then found 5559 keywords.   We ourselves then cleaned the resulting files.

Is this word Aguiar a person's name?  Please answer Yes or no. No need to provide an explanation.

Here is a list of 100 of the keywords which it found:

\rem{
\begin{multicols}{3}
 \begin{tabular}{l} 
effect\\
effective category\\
effective descent morphism\\
effective descent type\\
effective orbifold\\
effective quotient\\
effective topos\\
effective\\
effectiveness\\
effectivization\\
efficiently regular\\
electrical circuit\\
electrical engineering\\
element\\
elementary Seely category\\
elementary case\\
elementary category theory\\
elementary property\\
elementary quotient completion\\
elementary topos\\
eleutheric\\
embed\\
embeddable\\
embedded in cube\\
embedded\\
embedding result\\
embedding theorem\\
embedding\\
empty graph\\
enchilada category\\
encoding structure\\
endo-1\\
endo-2-functor\\
endodistributor\\
endofunction\\
endofunctor\\
endomodule\\
endomorphism monoid\\
endomorphism ring object\\
endomorphism ring\\
endomorphism\\
enlarged coherent site\\
enlarged site\\
enlargement\\
enriched 2-category\\
enriched Lawvere theory\\
enriched Yoneda lemma\\
enriched accessible category\\
enriched adjunction\\
enriched algebraic theory\\
enriched category theory\\
enriched category\\
enriched factorization system\\
enriched factorization\\
enriched functor\\
enriched graph\\
enriched icon\\
enriched indexed category\\
enriched model structure\\
enriched monad\\
enriched n\\
enriched orthogonality\\
enriched prefactorization system\\
enriched sheaf theory\\
enriched sketch\\
enriched\\
enrichment base\\
enrichment\\
entailment relation\\
entrance path category\\
entries-only\\
entropic Hopf algebra\\
entropic category\\
entropic setting\\
entropic version\\
entropy\\
entwining of functor\\
entwining operator\\
enumerative combinatoric\\
envelope\\
enveloping monoidale\\
epicomplete subcategory\\
epicomplete\\
epimorphic cover\\
epimorphic extension\\
epimorphic functor\\
epimorphic hull\\
epimorphic property\\
epimorphic\\
epimorphism\\    
epireflective protolocalisation\\
epireflective subcategory\\
epsilon-product\\
equaliser\\
equality of diagram\\
equality of map\\
equality of morphism\\
equality\\
equalizer\\
equation\\
equational axiom\\
\end{tabular}
\end{multicols}

\rem{
equational category
equational characterization
equational consequence
equational hull
equational presentation
equational structure
equational subcategory
equational theory
equationally presentable
equipment
equivalence class
equivalence of category
equivalence of proof
equivalence relation
equivalence theorem
equivalence
equivalent condition
equivalent effect
equivalent formulation
equivalent property
equivalent
equivariant A-category
equivariant Brauer group
equivariant Lusternik-Schnirelmann category
equivariant Morita equivalence
equivariant Serre-Swan theorem
equivariant algebraic K-homology
equivariant cohomology
equivariant operad
equivariant principal bundle
equivariant sheaf
equivariant simplicially enriched operadic world
equivariant stable homotopy theory
equivariant topological field theory
equivariant
ergodic action
erratic choice
esquisse
essential localization
essential new notion
essential
essentially affine category
essentially algebraic theory
essentially surjective functor
essentially surjective
etale groupoid
etendue
evaluation
eventually cyclic Boolean flow
eventually cyclic spectrum
exact Mal\'cev category
exact Maltsev category
exact category
exact completion
exact endofunctor
exact functor
exact homological category
exact horn-lifting
exact morphism
exact protomodular category
exact sequence
exact
exactness condition
exactness property
exactness
example
exchangable-output
execution path
existence of composite
existence problem
existence result
existence theorem
existence
existential quantification
existentially closed embedding
expectation
explicit construction
explicit description
explicitly defined
exponentiability
exponentiable morphism
exponentiable object
exponentiable space
exponentiable
exponential functor
exponential principle
exponential
exponentiation of locale
expressive
extended Artin-Mazur codiagonal
extended algebraic geometry
extended theory
extension of complex
extension system
extension
extensionality
extensive category
extensive quantity
extensive topos doctrine
extensive
exterior derivative
exterior differentiation
external Hom
external behavior
external observation
extra structure
extranatural transformation
extraspecial commutative Frobenius monoid
extremal disconnectedness
extremal functor
extremally disconnected
extriangulated category
}

}

ChatGPT didn't know that C-system

\section{The TAC Corpus and the 3000 Extracted Sentences}

Here is an example of an abstract of a paper that is found in the corpus:

\begin{quote}
Title: Exact sequences in the enchilada category.

Abstract:
We define exact sequences in the enchilada category of
C*-algebras and correspondences, and prove that the
reduced-crossed-product functor is not exact for the enchilada
categories. Our motivation was to determine whether we can have
a better understanding of the Baum-Connes conjecture by using
enchilada categories. Along the way we prove numerous results
showing that the enchilada category is rather strange
\end{quote}

\subsection{Extracted sentences}

Here are some of the sentences which were extracted, together with

{"context": "The general goal of this paper is to find the \"cyclic parts\" of such a flow", "concepts": ["'cyclic parts'", "'flow'"]}

{"context": "To do this, we approximate (X,t) by a flow on a Stone space (that is, a totally disconnected, compact Hausdorff space)", "concepts": ["'approximate'", "'flow'", "'Stone space'", "'totally disconnected'", "'compact Hausdorff space'"]}

{"context": "Such a flow can be examined by analyzing the resulting flow on the Boolean algebra of clopen subsets, using the spectrum defined in our previous TAC paper, The cyclic spectrum of a Boolean flow", "concepts": ["'flow'", "'Boolean algebra'", "'clopen subset'", "'spectrum'"]}

{"context": " In this paper, we describe the cyclic spectrum in terms that do not rely on topos theory", "concepts": ["'cyclic spectrum'", "'topos theory'"]}

{"context": "We then compute the cyclic spectrum of any finitely generated Boolean flow", "concepts": ["'cyclic spectrum'", "'finitely generated Boolean flow'"]}

{"context": "We define when a sheaf of Boolean flows can be regarded as cyclic and find necessary conditions for representing a Boolean flow using the global sections of such a sheaf", "concepts": ["'sheaf of Boolean flows'", "'cyclic sheaf'", "'Boolean flow'", "'global sections'", "'necessary conditions'"]}

{"context": "In the final section, we define and explore a related spectrum based on minimal subflows of Stone spaces.", "concepts": ["'minimal subflow'", "'Stone space'", "'spectrum'"]}

{"context": "The correlators of two-dimensional rational conformal field theories that are obtained in the TFT construction of Fuchs, Runkel and Schweigert are shown to be invariant under the action of the relative modular group and to obey bulk and boundary factorisation constraints", "concepts": ["'conformal field theory'", "'TFT construction'", "'correlator'", "'relative modular group'", "'bulk factorisation constraint'", "'boundary factorisation constraint'"]}

{"context": "We present results both for conformal field theories defined on oriented surfaces and for theories defined on unoriented surfaces", "concepts": ["'conformal field theory'", "'oriented surface'", "'unoriented surface'"]}

{"context": "In the latter case, in particular the so-called cross cap constraint is included.", "concepts": ["'cross cap constraint'"]}

\section{Pipeline for human annotators}

We carried out the following steps with  300 sentences
from the TAC corpus
 that three people from our team worked on.

\begin{enumerate}
\item 
We asked each person to read the 300 sentences and then to list the mathematical keywords.
\item
For each pair $(a,b)$ of different people from the three annotators on our team, we kept track of the keywords
listed by $a$ but not by $b$.

\end{enumerate}

\section{Pipeline for ChatGPT}

We carried out the following steps with both the 3000 sentences from TAC,
and also the smaller 300 sentences that three people from our team worked on.

\begin{enumerate}
\item We asked ChatGPT to extract mathematical concepts.  It found close to 6,800.

\begin{quote}
"""Given the following Context, extract the words that denote Math concepts. 
                                            Here are some examples:
                                            {ices}
                                            Also note that we are looking for concepts like \verb |{positive_words}|, but not words shown in daily English sentences like \verb |{negative_words}| !
                                            Moreover, please make the concept words singular. For example when we see 'functors' in a sentence, we would extract 'functor' rather than 'functors'!
                                            Now please solve the following problem.

                                            Context: \verb |{context} |

                                            Concepts:  ["""
\end{quote}

\item We ran the results above through a heuristic filter.  This eliminated concepts formulated in the plural,
removing LaTeX terms, numbers, items that consisted of single words (?), person's names recognized as such by spaCy.

We also removed the word "none" and similar words. About 1,200 of these were found, leaving 5,600.
That is, we removed the responses where ChatGPT said that it found nothing and gave a short explanation of that fact.

\label{filter}
\item We finally asked ChatGPT to classify the remaining concepts as ``mathematical'' or not.  
We did this by asking, for example

\begin{quote}
In the sentence \verb |'{example["sentence"]}'|, is the word \verb |{context}| a math concept? Please answer Yes or No? No need to provide an explanation.
\end{quote}

`Is this word Aguiar a person's name?  Please answer Yes or no. No need to provide an explanation.''
ChatGPT found 1551 expressions which it claimed were \emph{not} mathematical concepts, such as ``Aguiar'' just above.

However, many of the concepts which it said were mathematical, really weren't.

\item 
\label{filter-with-context}
Then we did the same thing with the context appended, in order to increase the accuracy.

In the sentence ?In the particular example when M is an appropriate 2-subcategory of Cat, this yields a conceptually different proof of some recent results due to Aguiar, Haim and Lopez Franco.?, is the word Aguiar a math concept? Please answer Yes or No? No need to provide an explanation.

This left 200 for removal.

\label{classify}
\item We removed the 200 items found in Step~\ref{classify} from the list of 5,400 items found after Step~\ref{filter}.

\end{enumerate}

\subsection{Head-to-head comparison of three annotators on 300 sentences}

\subsection{Running the ChatGPT on the pipeline for the 300 sentences considered by humans}

ChatGPT at first provided a list of 455 concepts for us, and we rejected 30.  The rejects are shown below.
Again, these were the `No' answers from the very last prompt (Step~\ref{filter-with-context}):

\begin{center}
\begin{tabular}{l}
1989\\
A. Kock\\
KZ-doctrine\\
M. Barr\\
application\\
building\\
characterisation\\
composition\\
doctrinal setting\\
example\\
logical development\\
\end{tabular}
\quad
\begin{tabular}{l}
machine behavior\\
persistence \\
sober\\
static program analysis\\
step\\
symbolic structure\\
time\\
type\\
approach\\
as there is no mention of \\
\qquad any math concepts in the context\\
\end{tabular}
\quad
\begin{tabular}{l}
branching\\
comparison\\
consequence\\
generalisation\\
literature\\
misbehavior\\
object\\
positive answer\\
unpublished work\\
\\
\\
\end{tabular}
\end{center}

\subsection{Comparing ChatGPT's output to the human-generated ``gold'' standard}

At the end, we compared ChatGPT's 455 keywords to those obtained from a composite list of 475 coming from the three annotators.
Here is a Venn diagram showing the numbers

 \def\firstcircle{(0,0) circle (1.5cm)}
\def\secondcircle{(0:2cm) circle (1.5cm)}

\colorlet{circle edge}{blue!50}
\colorlet{circle area}{blue!20}

\tikzset{filled/.style={fill=circle area, draw=circle edge, thick},
    outline/.style={draw=circle edge, thick}}

\setlength{\parskip}{5mm}
\[
\begin{tikzpicture}
    \begin{scope}
        \clip \firstcircle;
        \fill[filled] \secondcircle;
    \end{scope}
    \draw[outline] \firstcircle node {$80$};
    \draw[outline] \secondcircle node {$100$};
\end{tikzpicture}
\]
The intersection consists of 375 concepts.  As you can see there were 80 concepts identified by ChatGPT but not by 
our annotators, and $100$ concepts the other way.

\rem{
  \def\firstcircle{(90:1.75cm) circle (1.5cm)}
  \def\secondcircle{(210:1.75cm) circle (2.5cm)}
  \def\thirdcircle{(330:1.75cm) circle (2.5cm)}
  \[
    \begin{tikzpicture}
      \begin{scope}
    \clip \secondcircle;
    \fill[green] \thirdcircle;
      \end{scope}
      \begin{scope}
    \clip \firstcircle;
    \fill[cyan] \thirdcircle;
      \end{scope}
      \draw \secondcircle node [text=black,below left] {$B$};
      \draw \thirdcircle node [text=black,below right] {$C$};
    \end{tikzpicture}
    \]
    }


}
\bibliography{references}

\clearpage
\appendix
\section{Annotation platform}

To help human annotators create eventual golden standards for the extraction of mathematical concepts, we repurposed a tool
from the Natural Language Inference (NLI) field.
The tool shows annotators a sentence (possibly) containing mathematical concepts, and annotators will cut-and-paste spans of text that they believe are mathematical concepts. If the annotator believes there are no mathematical concepts, they can simply skip to the next sentence.
Annotators can also modify the span: they can change plurals to singluras,
or add or delete words, etc.   

Development of this tool is ongoing.  But it can be used as-is with other experiments.
It can compare the performance of human annotators or between humans and LLMs.
 We also have the possibility of ``adjudicating'' 
 disagreements
 between annotations. In the near future we hope to
expand the tool to add crowd-source interface, in order to 
 allow mathematicians  to ``correct" machine annotations and to express opinions
 on tricky cases.

\label{sec:platform}

\begin{figure*}[h]
\begin{center}
\includegraphics[width=\textwidth]{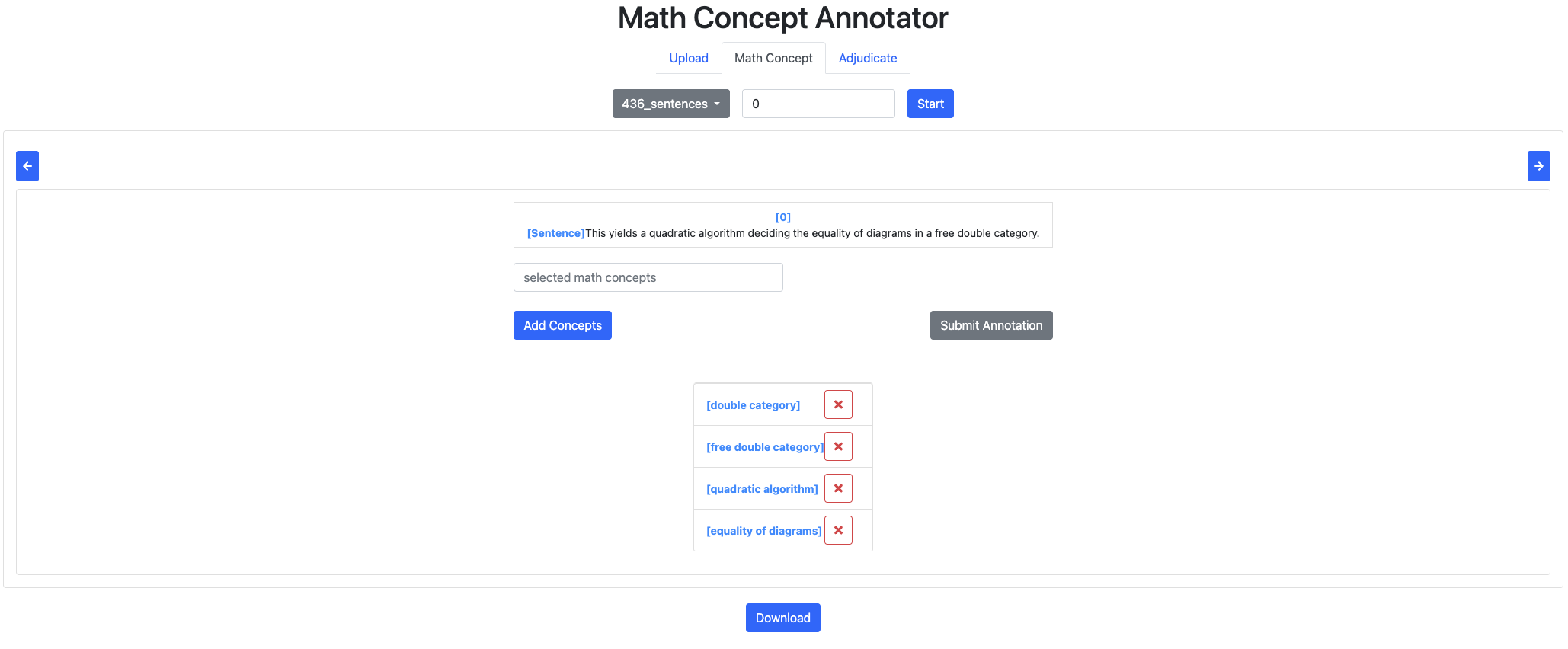}
\includegraphics[width=\textwidth]{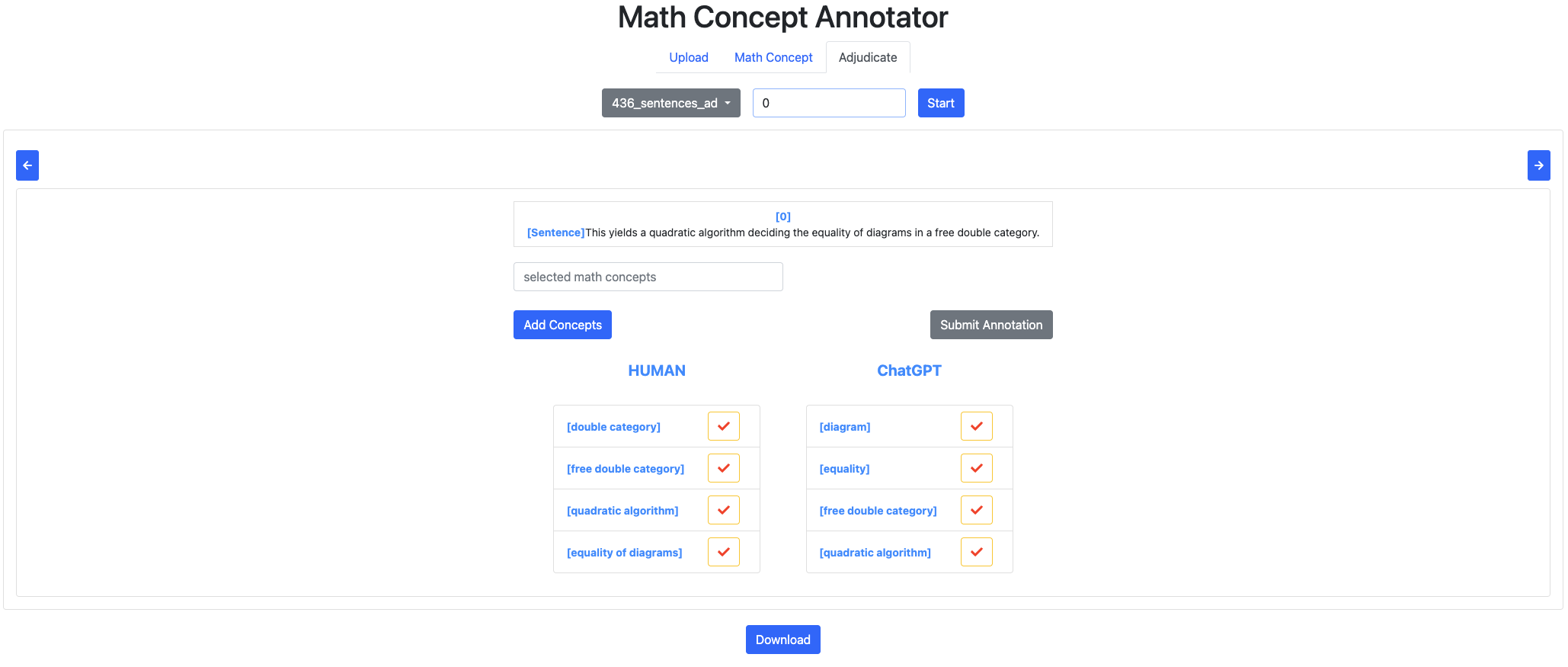}
\end{center}
\caption{Overview of the annotation platform when an input sentence is selected.}
\label{fig:platform_overview}
\end{figure*}

The tool is availalble at \url{https://gaoq111.github.io/math_concept_annotation/}.

\end{document}